\documentclass{article}

\usepackage[final]{corl_2026}         %

\title{Push the Limit: Learning High-Speed Sprinting and Aggressive Cornering on a Full-Scale Humanoid}

\author{
  Huaxing Huang\textsuperscript{1}\thanks{Equal contribution.} \And
  Ruidong Li\textsuperscript{2}\footnotemark[1] \And
  Jinchao Han\textsuperscript{1} \And
  Ziyang Dong\textsuperscript{1}
\end{tabular}
\begin{tabular}[t]{c}\bf\rule{0pt}{16pt}%
  Wenhao Cui\textsuperscript{1}\thanks{Corresponding author.} \\[0.5em]
  \textsuperscript{1}Noetix Robotics \quad
  \textsuperscript{2}Tsinghua University \\
  \texttt{wenhaocui@noetixrobotics.com}
}

\usepackage{xspace}     %
\usepackage{booktabs}   %
\usepackage{amsmath}    %
\usepackage{amssymb}    %
\usepackage{float}      %
\usepackage{enumitem}   %
\usepackage{wrapfig}    %
\usepackage{graphicx}   %
\usepackage{subcaption}  %
\graphicspath{{figures/}}

\newcommand{\vx}{v_x}                               %
\newcommand{\vy}{v_y}                               %
\newcommand{\cmdvel}{\mathbf{c}}                    %
\newcommand{\obs}{\mathbf{o}}                       %
\newcommand{\act}{\mathbf{a}}                       %
\newcommand{\skillvec}{\mathbf{g}}                  %

\begin{document}
\maketitle

\vspace{-1.9em}
\begin{figure*}[h]
\centering
\includegraphics[width=0.92\textwidth]{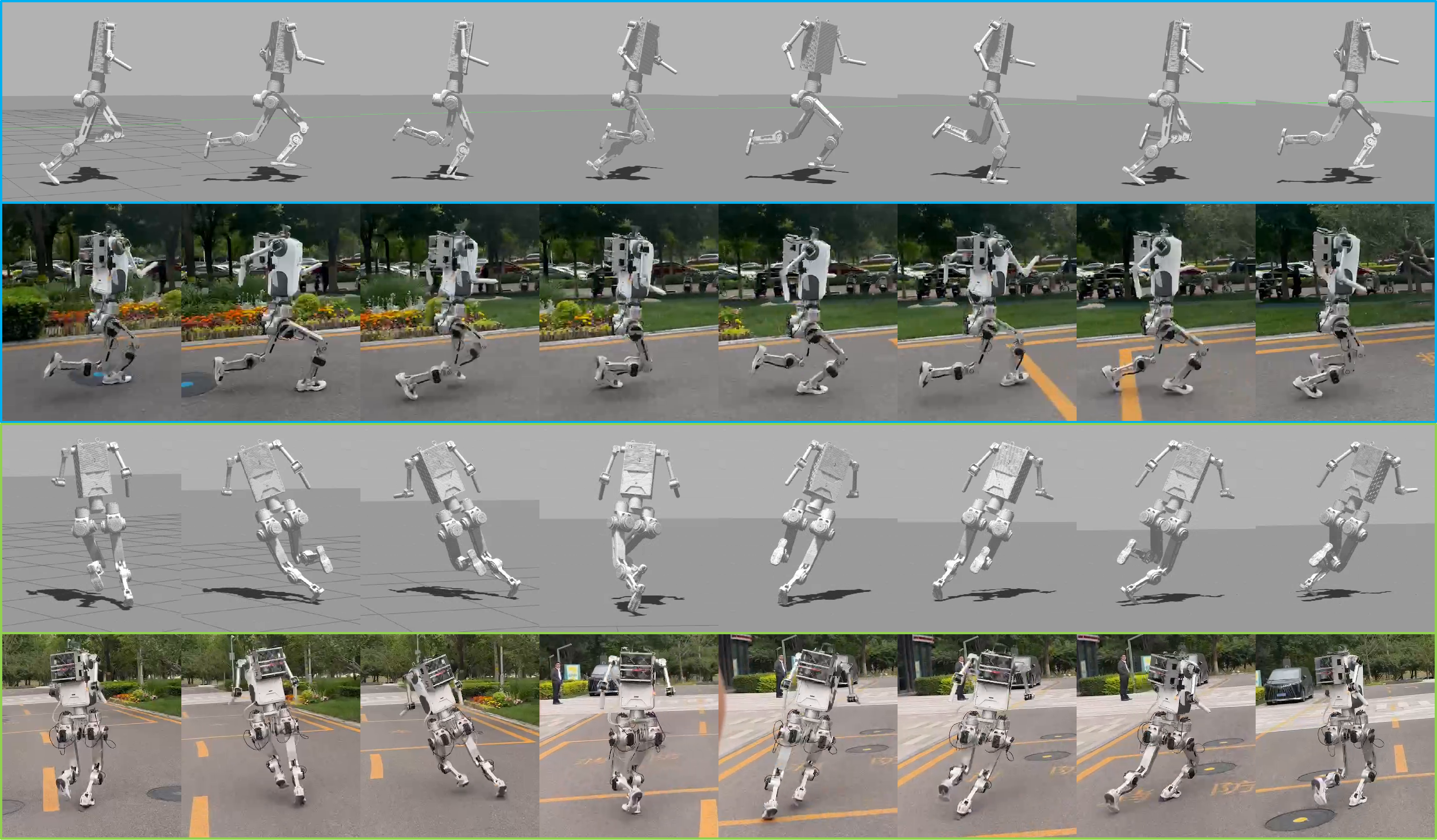}%
\caption{%
Gait sequences from simulation (Gazebo, rows 1 \& 3) and real-robot deployment (rows 2 \& 4).
\textbf{Rows 1--2:} high-speed sprinting with motion-prior-guided stride dynamics at velocity command $6.5\,\mathrm{m/s}$ (timed hardware speed reported in Sec.~\ref{sec:exp:hardware}).
\textbf{Rows 3--4:} aggressive cornering with maximum trunk lean exceeding $25^\circ$ at $5.5\,\mathrm{m/s}$, an emergent inward-leaning strategy that arises from the heading-direction reward without explicit posture supervision.
}
\label{fig:overview}
\vspace{-0.3em}
\end{figure*}

\begin{abstract}
We present a learning-based framework that enables a full-scale bipedal humanoid to sprint at over $7\,\mathrm{m/s}$ with motion-prior-guided gait and perform aggressive cornering at high speed with a pronounced trunk lean of up to $25^\circ$. Achieving this combination is challenging: above approximately 4 m/s, periodic foot impacts induce severe, high-frequency yaw-rate oscillations intrinsic to high-speed gaits. This high-frequency temporal structure corrupts the history of proprioceptive feedback that flat-MLP policies rely on, while rendering conventional yaw-rate tracking rewards counterproductive, as they fight against the robot's natural oscillatory dynamics. Moreover, a single style prior shared across skills cannot tell which skill is commanded and inherently compromises the low-speed skill fidelity. To overcome these limitations, we adopt a causal multi-head attention (MHA) module to explicitly model the high-frequency temporal feedback. In place of the oscillatory yaw rate, we design a heading-direction reward that supervises only the low-frequency velocity direction, allowing agile turning at high speed without suppressing natural gait oscillations. We further introduce skill-conditioned multi-adversarial-motion-prior (MAMP) discriminators that condition the style reward on the commanded skill. Ablation experiments confirm that each mechanism is essential: replacing structured temporal encoding  with a flat-history MLP caps the top speed; reverting to yaw-rate tracking prevents high-speed turning; and a single shared discriminator makes in-place turning collapse into a quasi-static stance. Our framework bridges the gap between raw sprinting speed and reference-motion fidelity, demonstrating stable high-speed sprinting and aggressive cornering in both simulation and real-robot deployment (videos: \url{https://youtu.be/YWf62XkiApc}).
\end{abstract}


\section{Introduction}
\label{sec:intro}

Bipedal robots can now sprint in a straight line at speeds exceeding those of human athletes, yet the resulting gaits bear little resemblance to human running---arms flail, torsos remain rigidly upright, and high-speed turning is absent from published demonstrations.
Conversely, learning-based controllers that leverage adversarial motion priors produce convincingly human-like movement, but only at moderate speeds and with limited agility~\citep{peng2021amp, escontrela2022adversarial, radosavovic2024real}.
The combination that matters for real-world deployment---sustained sprinting with natural stride dynamics and agile cornering---remains undemonstrated on physical hardware.
Achieving this combination is a system-level stress test that simultaneously exposes weaknesses in state estimation, motion-style supervision, and reward design, weaknesses that remain hidden at walking pace.

Three specific failure modes emerge when a learned bipedal controller is pushed into the high-speed regime.
\textbf{(i)}~Above approximately $4\,\mathrm{m/s}$, the proprioceptive stream is dominated by high-frequency oscillations from alternating leg impacts; a standard MLP consuming a flat observation history cannot extract the temporal structure---relative phase, contact timing, acceleration profiles---needed for stable state estimation at these frequencies.
\textbf{(ii)}~The conventional yaw-rate tracking reward breaks down at sprinting speeds: alternating leg strides force the torso to oscillate in yaw, making instantaneous yaw rate a noisy mixture of the true turning signal and gait-induced oscillation. The policy either slows down to produce a cleaner signal or abandons the turning objective altogether, failing to achieve effective yaw-rate command tracking at sprinting speeds.
\textbf{(iii)}~When multiple locomotion skills (standing, turning, walking, sprinting) share a single adversarial discriminator trained on the mixed reference dataset, the discriminator rates whether a motion resembles any reference skill, not whether it matches the commanded one. A skill that the task reward barely distinguishes from standing then receives almost no learning signal; in our setting the policy loses the ability to turn in place---a mode collapse across the skill space.

Here we present a learning-based locomotion control framework that addresses each failure mode through a targeted design choice:
(i)~a \emph{causal multi-head attention (MHA) history encoder} that replaces the flat-history MLP, extracting a compact dynamics representation from the proprioceptive window without increasing network capacity;
(ii)~a \emph{heading-direction reward} that supervises the low-frequency velocity direction rather than the high-frequency instantaneous yaw rate, restoring a usable turning signal at full sprint; and
(iii)~a \emph{skill-conditioned Multi-AMP} framework whose per-skill discriminators condition the style reward on the commanded skill, preventing mode collapse across the skill set.
The resulting policy is deployed zero-shot on a physical $17$-DOF bipedal humanoid; quantitative evaluation in Gazebo simulation confirms $6.5\,\mathrm{m/s}$ velocity tracking and aggressive cornering at high speed (Figure~\ref{fig:overview},\ref{fig:turning_e2}), while a timed $100\,\mathrm{m}$ hardware sprint confirms an average speed of $7.14\,\mathrm{m/s}$ (Sec.~\ref{sec:exp:hardware}). Ablation experiments isolate each component: structured temporal encoding is necessary for high-speed tracking, and the MHA encoder further enables stable aggressive cornering where GRU, LSTM, and learned-estimator alternatives fail (E1); replacing heading-direction with yaw-rate tracking eliminates agile high-speed turning (E2); and a single shared discriminator causes in-place turning to collapse (E3). We validate all claims through standardized step-speed and turning test protocols in both simulation and real-robot deployment.

\section{Related Work}
\label{sec:related}

\textbf{Adversarial motion priors.}
AMP~\citep{peng2021amp} introduced a GAIL-style discriminator that provides a unified style reward from unstructured motion clips, removing the need for explicit clip selection required by tracking-based methods such as DeepMimic~\citep{peng2018deepmimic}. WASABI~\citep{li2022wasabi} adopted a Wasserstein formulation to learn agile skills from rough, partial demonstrations, and HumanMimic~\citep{tang2024humanmimic} applied Wasserstein adversarial imitation to natural humanoid locomotion and gait transitions. Multi-AMP~\citep{vollenweider2022multiamp} extends this to one discriminator per motion style for quadrupeds, so that each skill receives its own style reward, but has not been applied to regimes spanning standing through high-speed sprinting with policy-side skill conditioning.

\textbf{Attention-based history encoders.}
Most locomotion controllers consume flat proprioceptive histories via an MLP~\citep{rudin2022learning, kumar2021rma} or decouple history processing into an auxiliary estimator~\citep{long2024hybrid}. More broadly, Transformer architectures have shown promise for sequential decision-making in RL~\citep{chen2021decision} and embodiment-aware policy learning~\citep{sferrazza2024body}, motivating their adoption in locomotion control. \citet{ma2026rgmt} demonstrated that causal self-attention over proprioceptive history substantially improves humanoid motion tracking, using cross-attention to fuse high-dimensional reference trajectories.

\textbf{High-speed bipedal locomotion.}
Sim-to-real RL for bipeds has progressed from periodic gait composition~\citep{siekmann2021gaits} to versatile dynamic control encompassing walking, running, and jumping on Cassie~\citep{li2024reinforcement}. On the quadruped side, \citet{margolis2022rapid} demonstrated $3.9\,\mathrm{m/s}$ sprinting via model-free RL. For humanoids, \citet{radosavovic2024real} achieved robust outdoor walking with a causal Transformer policy, and denoising world-model learning extended blind locomotion to challenging terrains~\citep{gu2024dwl}. Vision-based parkour policies traverse obstacles such as platforms, hurdles and gaps~\citep{zhuang2024parkour}, and unified whole-body controllers command versatile gaits with fine-grained parameters such as step frequency and foot height~\citep{xue2025hugwbc}. In parallel, \citet{cheng2024expressive} demonstrated expressive whole-body motion tracking, \citet{fu2024humanplus} enabled humanoid teleoperation through human motion shadowing, ASAP~\citep{he2025asap} reduced the sim-to-real dynamics gap for agile whole-body skills, and \citet{liao2024berkeley} introduced a lightweight humanoid research platform for learning-based control. These works target either moderate speed with high robustness, or full-body motion tracking at low to moderate speeds; none addresses the coupled failure modes---temporal state estimation, turning reward corruption, and multi-skill style collapse---specific to sustained high-speed bipedal sprinting that we identify in Sec.~\ref{sec:intro}.

\begin{figure}[t]
\centering
\includegraphics[width=0.87 \textwidth]{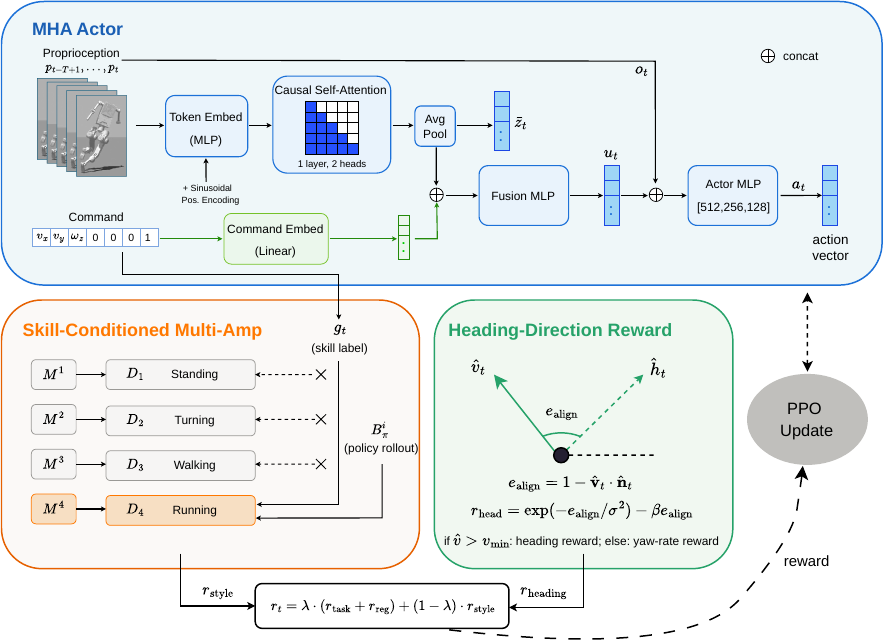}%
\caption{%
System overview. \textbf{(1)}~The MHA history encoder processes a $T$-frame proprioceptive window via causal self-attention and fuses the dynamics summary with the current velocity-plus-skill command to produce joint position targets. \textbf{(2)}~Skill-conditioned Multi-AMP discriminators provide an independent style reward per skill, conditioning the style reward on the commanded skill. \textbf{(3)}~The heading-direction reward supervises the low-frequency velocity direction at high speed, falling back to yaw-rate tracking below $v_{\min}$. All three components feed into a shared PPO training loop.
}
\label{fig:framework}
\end{figure}

\section{Method}
\label{sec:method}

We learn a single feedback controller that maps proprioceptive history and a velocity-plus-skill command to joint position targets, addressing the three failure modes identified in Sec.~\ref{sec:intro} through the MHA history encoder (Sec.~\ref{sec:method:mha}), heading-direction reward (Sec.~\ref{sec:method:yawhead}), and skill-conditioned Multi-AMP discriminator (Sec.~\ref{sec:method:camp}). Figure~\ref{fig:framework} shows the overall architecture; we begin with the problem formulation.

\subsection{Problem Formulation}
\label{sec:method:problem}

We model locomotion as a discrete-time Markov decision process $\mathcal{M} = (\mathcal{S}, \mathcal{A}, p, r, \gamma)$ at $50\,\mathrm{Hz}$ control rate. At step~$t$ the policy observes $\obs_t = (\cmdvel_t, \mathbf{p}_t)$, where $\cmdvel_t = [v^{\mathrm{cmd}}_x, v^{\mathrm{cmd}}_y, \omega^{\mathrm{cmd}}_z, \skillvec_t]\in\mathbb{R}^7$ packs the commanded base velocity together with a one-hot skill label $\skillvec_t\in\{0,1\}^4$ (\emph{standing}, \emph{in-place turning}, \emph{walking}, \emph{running}), and $\mathbf{p}_t\in\mathbb{R}^{56}$ collects base angular velocity, base Euler angles, joint positions, joint velocities and the previous action. The agent emits $\act_t\in\mathbb{R}^{17}$ as target joint angles; a low-level proportional-derivative controller produces the applied torques. The policy operates over a length-$T$ history $\mathbf{H}_t = (\obs_{t-T+1}, \ldots, \obs_t)$; flattening this history yields the full actor input $\mathbf{x}_t = \mathrm{vec}(\mathbf{H}_t)\in\mathbb{R}^{TD}$ with per-step dimension $D=|\obs_t|$.

The skill label~$\skillvec_t$ is generated automatically from~$\cmdvel_t$ inside the command term: \emph{running} is selected when $v^{\mathrm{cmd}}_x\ge 2.0\,\mathrm{m/s}$, \emph{in-place turning} when the linear command is near zero but $|\omega^{\mathrm{cmd}}_z|$ is appreciable, and \emph{standing} for vanishing commands; otherwise the label defaults to \emph{walking}. This deterministic mapping lets the controller share a single neural network across skills while still receiving an explicit categorical context.

The reward blends task, regularization, and style components as $r_t = \lambda\,(r^{\mathrm{task}}_t + r^{\mathrm{reg}}_t) + (1{-}\lambda)\,r^{\mathrm{style}}_t$ with $\lambda{=}0.8$, where $r^{\mathrm{task}}_t$ tracks the commanded $(\vx, \vy)$ and the heading-direction objective of Sec.~\ref{sec:method:yawhead}, $r^{\mathrm{style}}_t$ comes from the AMP discriminator(s) of Sec.~\ref{sec:method:camp}, and $r^{\mathrm{reg}}_t$ collects standard regularizers (joint-limit, torque-limit, action-rate and energy terms; full table in the appendix). The policy is optimized with PPO~\citep{schulman2017proximal} and generalized advantage estimation~\citep{schulman2015high}.

\subsection{Multi-Head Attention Actor}
\label{sec:method:mha}

To address the flat-history limitation described in Sec.~\ref{sec:intro}, we adopt the causal self-attention history encoder of~\citet{ma2026rgmt}, which extracts a compact dynamics representation from proprioceptive history via multi-head self-attention. We simplify the command interface: the cross-attention encoder in~\citet{ma2026rgmt} targets high-dimensional reference trajectories ($\mathbb{R}^{38}$ per step); our velocity-plus-skill command ($\mathbb{R}^7$) is low-dimensional, so we replace cross-attention with MLP fusion. The encoder processes $T$ proprioceptive frames through a single-layer, two-head causal Transformer, average-pools across time to produce a dynamics summary, and fuses it with the embedded command; the result is passed to the actor MLP to produce joint-position targets (Figure~\ref{fig:framework}; architectural details in Appendix~\ref{app:arch}).

The MHA actor totals $3.6\%$ fewer parameters than the MLP baseline consuming the same history window, so any performance gain reflects the structural inductive bias of temporal attention rather than added capacity. In practice, the MHA design yields substantially stronger speed generalization and high-speed turning stability than the MLP baseline (Sec.~\ref{sec:exp:e1}).

\subsection{Heading-Direction Reward}
\label{sec:method:yawhead}

As noted in Sec.~\ref{sec:intro}, the conventional yaw-rate tracking reward degrades at sprinting speeds because the torso must oscillate in yaw to counterbalance alternating leg strides, corrupting the instantaneous~$\omega_z$ signal. We instead supervise the direction of the base velocity against a maintained heading target. For every episode during training, we sample a target heading direction around the initial robot heading. Let $\mathbf{v}^{\mathrm{xy}}_t$ denote the world-frame planar base velocity and $\psi_t^{\star}$ the commanded heading. Define $\hat{\mathbf{v}}_t = \mathbf{v}^{\mathrm{xy}}_t / \|\mathbf{v}^{\mathrm{xy}}_t\|$ and $\hat{\mathbf{n}}_t = (\cos\psi_t^{\star}, \sin\psi_t^{\star})$. We adopt a heading-direction reward
\begin{equation}
\label{eq:heading_reward}
r^{\mathrm{head}}_t = \exp\!\Big(\!-\frac{e_{\text{align}}}{\sigma^2}\Big) - \beta\,e_{\text{align}},
\qquad
e_{\text{align}}=1-\hat{\mathbf{v}}_t\!\cdot\!\hat{\mathbf{n}}_t,
\end{equation}
when the commanded linear speed exceeds a threshold $v_{\min}=2.0\,\mathrm{m/s}$, and we otherwise fall back to the conventional yaw-rate term
\begin{equation}
\label{eq:yawrate_reward}
r_t^{\omega}=\exp\!\left(-\frac{e_{\omega_z}^2}{\sigma^2}\right)-\beta\,|e_{\omega_z}|,
\qquad
e_{\omega_z} = \omega_z - \omega^{\mathrm{cmd}}_z.
\end{equation}
The velocity direction changes far slower than the gait frequency because the base translational velocity is the time-integral of net forces divided by body mass; rapid oscillatory forces largely cancel over a full gait cycle, leaving only the low-frequency turning trend. The instantaneous yaw rate, by contrast, captures every rotational perturbation including gait-induced torso rocking. The bias term $-\beta\,e_{\text{align}}$ adds an L1 gradient outside the exponential's saturation regime, which is critical at the start of training when alignments are far from one.
Let $\psi_t$ denote the current base heading angle. The heading error is also used to derive a proportional commanded yaw-rate $\omega^{\mathrm{cmd}}_z$ during the training so that the velocity command remains consistent with the heading target:
\begin{equation}
\label{eq:yawrate_cmd}
    \omega^{\mathrm{cmd}}_z = \mathrm{clip}\!\big(k_\psi (\psi^{\star}_t - \psi_t),\; \omega_{\min},\; \omega_{\max}\big)
\end{equation}

\subsection{Skill-Conditioned Multi-AMP Discriminator}
\label{sec:method:camp}

To counter the mode collapse described in Sec.~\ref{sec:intro}, we follow~\citet{vollenweider2022multiamp} and instantiate $K{=}4$ skill-specific discriminators $\{D_i\}_{i=1}^{K}$, each with its own reference dataset~$\mathcal{R}^{i}$ and policy replay buffer~$\mathcal{B}^{i}_{\pi}$. For a transition emitted under active skill~$i$, only $D_i$ updates and only $D_i$ provides the style reward. This routing makes the style reward conditional on the commanded skill, so that producing the motion of another skill is penalized (Sec.~\ref{sec:exp:e3}).

The multi-discriminator structure follows~\citet{vollenweider2022multiamp}; we employ a Wasserstein GAN divergence for the discriminator loss, which replaces AMP's original least-squares loss with a Wasserstein distance that maintains informative gradients even when policy and reference distributions are far apart. We introduce three stabilizing modifications for the high-speed bipedal regime: (i)~a $\tanh$ output bound $\tanh(\alpha\,x)$ that prevents gradient explosion from unbounded discriminator scores; (ii)~a non-negative reward mapping $\rho(x) = \exp(\tanh(\alpha\,x))-e^{-1}$ that eliminates negative style rewards---without this, the policy learns to terminate episodes early to avoid accumulating negative style penalty; and (iii)~gradient penalties on both expert and policy samples ($R_1 + R_2$ regularization), which smooth the discriminator's decision boundary across the full motion manifold rather than only near the expert distribution. The resulting discriminator loss for style~$i$ is
\begin{equation}
\label{eq:disc_loss}
\mathcal{L}_{D_i} = \tfrac{1}{2}\!\Big(\!-\mathbb{E}_{\mathcal{R}^i}\!\big[\tanh\!\big(\alpha\,D_i(\boldsymbol{\phi})\big)\big] + \mathbb{E}_{\mathcal{B}^i_\pi}\!\big[\tanh\!\big(\alpha\,D_i(\boldsymbol{\phi})\big)\big]\Big) + \mathbb{E}_{\mathcal{R}^i}\!\big[\|\nabla_{\boldsymbol{\phi}} D_i\|^p\big] + \mathbb{E}_{\mathcal{B}^i_\pi}\!\big[\|\nabla_{\boldsymbol{\phi}} D_i\|^p\big],
\end{equation}
where $\boldsymbol{\phi}$ denotes the AMP observation window (composition detailed in Appendix~\ref{app:arch}). The style reward at step~$t$ is
\begin{equation}
\label{eq:style_reward}
r^{\mathrm{style}}_t = w_s\, \rho\big(D_{i(t)}(\boldsymbol{\phi}_t)\big),
\qquad
i(t) = \arg\max_i \skillvec_t[i],
\end{equation}
The full per-step reward is $r_t = \lambda\,(r^{\mathrm{task}}_t + r^{\mathrm{reg}}_t) + (1{-}\lambda)\,r^{\mathrm{style}}_t$ with $\lambda{=}0.8$, so that style supervision regularizes rather than overrides the velocity objective.

\subsection{Training and Deployment Pipeline}
\label{sec:method:pipeline}

We train in IsaacLab~\citep{mittal2023orbit} at $f_{\mathrm{sim}}\,\mathrm{Hz}$ physics with a control decimation of~$n_{\mathrm{dec}}$ ($f_{\mathrm{ctrl}}=f_{\mathrm{sim}}/n_{\mathrm{dec}}$), $N_{\mathrm{env}}$ parallel environments and episodes of $T_{\mathrm{ep}}\,\mathrm{s}$, on top of the massively parallel PPO implementation of~\citet{rudin2022learning}. The $K{=}4$ reference motion datasets total roughly $N_{\mathrm{ref}}$ pre-loaded transitions at $f_{\mathrm{ctrl}}\,\mathrm{Hz}$ playback rate; reference state initialization is applied to $20\%$ of resets to keep the high-speed regime visited from start. All numerical values are listed in Table~\ref{tab:hyperparams}.

Standard domain randomization~\citep{tobin2017domain, peng2018sim} is applied to friction, actuator parameters, base mass, and COM offset; a delayed implicit actuator model captures the first-order motor response. Full randomization ranges are listed in Appendix~\ref{app:sim2real}.

\textbf{Real-robot deployment.}
Trained policies are exported to ONNX and deployed zero-shot---without any real-world fine-tuning---on both Gazebo and the physical robot at $50\,\mathrm{Hz}$, using the same observation vector and PD gains as in simulation.

\section{Experiments}
\label{sec:experiments}

We evaluate the full pipeline and each of its three components in simulation and validate the full pipeline through real-robot deployment on a $17$-DOF bipedal humanoid. All policies share the same domain-randomization and training hyperparameters described in Sec.~\ref{sec:method:pipeline}; only the ablated component differs.

\subsection{Experimental Setup}
\label{sec:exp:setup}

The robot is a $17$-DOF bipedal humanoid ($10$ leg, $1$ waist, $6$ arm joints) standing $1.8\,\mathrm{m}$ tall, driven by brushless DC motors with $330\,\mathrm{Nm}$ peak torque and $19.15\,\mathrm{rad/s}$ no-load speed at the leg joints. The high-speed tests (E1, E2) run in Gazebo, which executes the same ONNX policy and ROS controller as the physical robot; the per-skill style analysis (E3) runs in Isaac Lab, where the training reward and all discriminators can be evaluated on the same rollouts. Real-robot experiments run the ONNX policy zero-shot inside the ROS controller at $f_{\mathrm{ctrl}}\,\mathrm{Hz}$.

As an external baseline we adopt the HIM+MLP architecture of~\citet{huang2025thinkfeetseamlesstransition}, retrained from scratch within our full pipeline (heading reward, Multi-AMP, speed curriculum) with matched parameter count and identical configuration.
We further compare the full model against three single-component ablations and two encoder variants:
\begin{itemize}[noitemsep, leftmargin=1.2em,topsep=0.5ex]
  \item \textbf{E1\,---\,Encoder comparison:} the causal Transformer is replaced with (a)~a flat-history MLP, (b)~a GRU encoder, or (c)~an LSTM encoder, each consuming the same history window with comparable parameter count. The HIM baseline also participates in this comparison.
  \item \textbf{E2\,---\,Yaw-rate reward:} the speed-dependent reward switching is removed; the yaw-rate reward (Eq.~\ref{eq:yawrate_reward}) is used at all speeds instead of switching to the heading-direction reward (Eq.~\ref{eq:heading_reward}) above $v_{\min}$.
  \item \textbf{E3\,---\,Shared discriminator:} the four per-skill discriminators are collapsed into a single shared discriminator trained on the mixed reference dataset.
\end{itemize}

We define two standardized test protocols executed with fixed command sequences (no joystick input), plus a motion-quality metric:
\begin{itemize}[noitemsep, leftmargin=1.2em,topsep=0.5ex]
  \item \textbf{Step test:} a square-wave velocity command that transitions from $(0,0,0)$ to $(v_{\max},0,0)$ with $v_{\max}=6.5\,\mathrm{m/s}$. We record the commanded and actual base-frame $v_x$ over time, measuring acceleration profile, steady-state tracking error, and maximum sustained speed.
  \item \textbf{Turning test:} a constant command $(v_x, 0, \omega_z)$ that asks the robot to turn while sprinting; the specific $(v_x, \omega_z)$ pair is stated with each result. We record the world-frame XY trajectory, trunk lean angle, turning radius, and multi-lap trajectory overlap.
  \item \textbf{DTW distance:} following WASABI~\citep{li2022wasabi}, we align 3-s segments of the per-frame AMP observation from steady-state rollouts with 6-s segments of the commanded skill's reference motion (L2 local cost, Mori asymmetric step pattern, open-ended matching) and report the mean per-frame distance along the alignment, measuring per-skill motion quality.
\end{itemize}

Finally, we validate the top-speed performance of our method on hardware via a timed $100\,\mathrm{m}$ sprint on a standard athletics track (Sec.~\ref{sec:exp:hardware}).
\subsection{Main Results}
\label{sec:exp:main}

In simulation, the full pipeline tracks the $6.5\,\mathrm{m/s}$ step command with a steady-state error of approximately $0.1\,\mathrm{m/s}$ and sustains stable cornering at full speed. The trunk lean angle during cornering depends on the centripetal acceleration $a_c = v \cdot |\omega_z|$: at $v{=}5.5\,\mathrm{m/s}$ with a larger commanded yaw rate, the policy achieves a maximum lean exceeding $25^\circ$ (confirmed in both simulation and real-robot deployment; Figure~\ref{fig:overview}); at $v{=}6.5\,\mathrm{m/s}$ with $\omega_z{=}0.35\,\mathrm{rad/s}$ (the standardized turning test, Figure~\ref{fig:turning_e2}), the lean reaches $13^\circ$ with a turning radius of $24.2\,\mathrm{m}$. Both regimes exhibit the emergent trunk-lean behavior characteristic of the heading-direction reward. On hardware, a timed $100\,\mathrm{m}$ sprint yields $7.14\,\mathrm{m/s}$ average speed (Sec.~\ref{sec:exp:hardware}). All ablation comparisons below are simulation results under identical conditions per pair of policies.

\subsection{Effect of History Encoder Architecture (E1)}
\label{sec:exp:e1}
\begin{figure*}[t]
\centering
\includegraphics[width=0.75\textwidth]{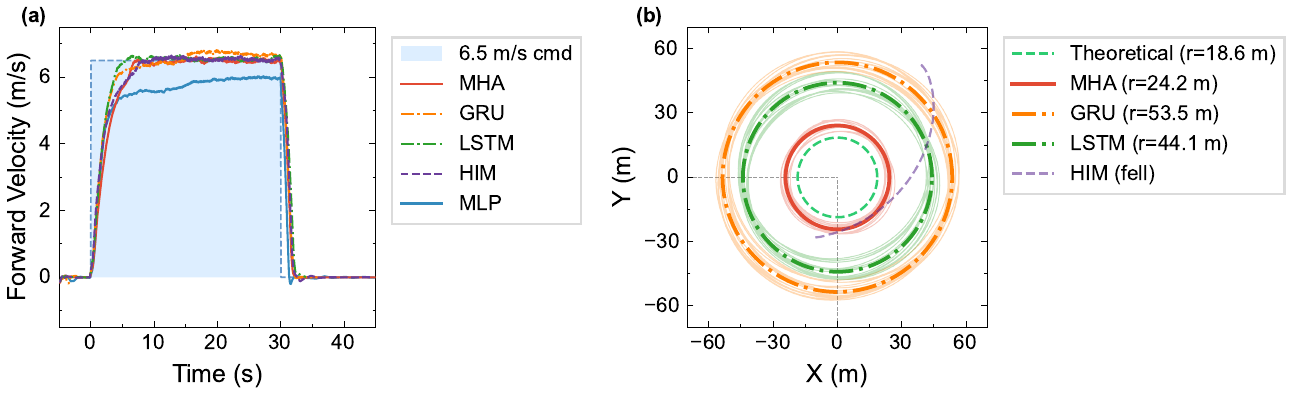}
\caption{%
Encoder comparison in Gazebo (E1).
\textbf{(a)}~Step-test velocity tracking: MHA, GRU, LSTM, and HIM all track the $6.5\,\mathrm{m/s}$ command, generalizing beyond the $5.5\,\mathrm{m/s}$ training curriculum; the flat-history MLP saturates near $6.0\,\mathrm{m/s}$. Curves smoothed with a $151$-point moving average.
\textbf{(b)}~Cornering trajectories under command $(6.5,\,0,\,0.35)$: MHA achieves a $24.2\,\mathrm{m}$ turning radius; LSTM and GRU require $1.8{\times}$/$2.2{\times}$ larger radii; HIM falls. The green dashed circle marks the theoretical radius $r{=}v/|\omega|{=}18.6\,\mathrm{m}$.
}
\label{fig:encoder_comparison}
\vspace{-0.6em}
\end{figure*}
All encoder variants are trained with the same velocity curriculum ($5.5\,\mathrm{m/s}$), reward, and Multi-AMP configuration; only the history encoder differs. On the $6.5\,\mathrm{m/s}$ step test (Figure~\ref{fig:encoder_comparison}a), MHA, GRU, LSTM, and HIM all track the target stably---generalizing $18\%$ beyond the curriculum limit---while the flat-history MLP saturates near $6.0\,\mathrm{m/s}$. This indicates that \textbf{high-speed straight-line tracking is enabled by structured temporal encoding in general}, rather than by any single architecture; the framework's reward design and speed curriculum are the primary contributors. A history-length sweep over $5$--$20$ frames showed negligible performance variation.

However, the architectures diverge sharply in cornering (Figure~\ref{fig:encoder_comparison}b). Under a $(6.5,\,0,\,0.35)$ command, MHA achieves a turning radius of $24.2\,\mathrm{m}$ with pronounced trunk lean; LSTM and GRU require $1.8{\times}$ and $2.2{\times}$ larger radii, respectively; the HIM baseline falls outright. \textbf{MHA's advantage concentrates in dynamic stability during aggressive maneuvers}, where the self-attention mechanism can selectively weight the most informative time steps within a gait cycle to maintain balance under high centripetal loads. All actors have comparable parameter counts (the MHA actor is $3.6\%$ smaller than the MLP), confirming that performance differences stem from the inductive bias of temporal modeling rather than capacity.

\subsection{Effect of Heading-Direction Reward (E2)}
\label{sec:exp:e2}

\begin{figure}[t]
\centering
\includegraphics[width=0.9\columnwidth]{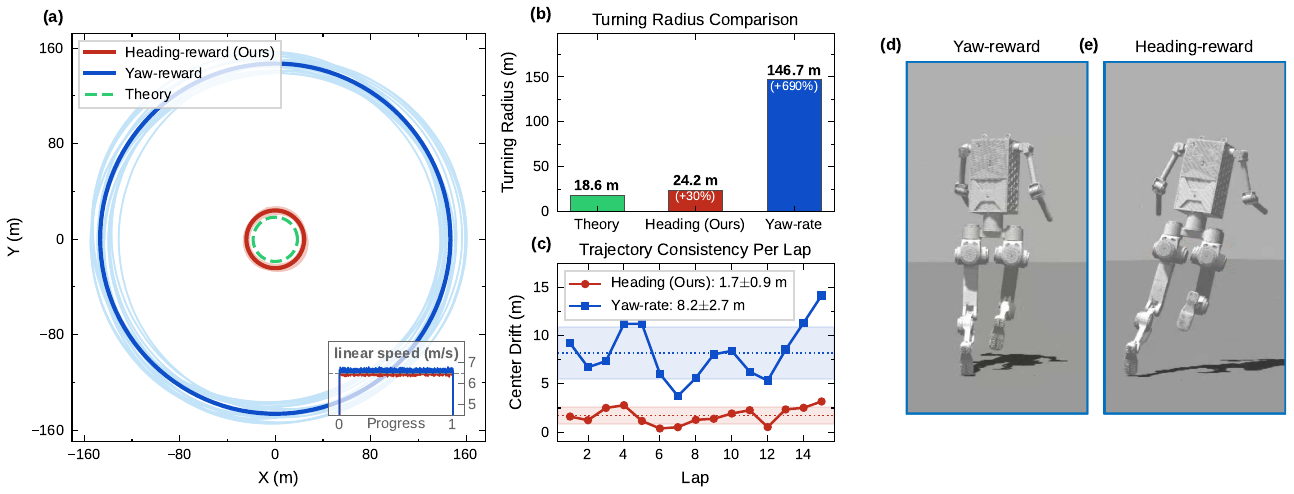}
\caption{%
Turning-test comparison in Gazebo (E2 ablation). Both the \emph{heading-reward policy} (ours, red) and the \emph{yaw-rate policy} (E2 baseline, blue) receive the same command $(v_x{=}6.5\,\mathrm{m/s},\;\omega_z{=}0.35\,\mathrm{rad/s})$; theoretical radius $r{=}v/|\omega|{=}18.6\,\mathrm{m}$.
\textbf{(a)}~World-frame XY trajectories over 15 laps. Thin transparent traces are raw trajectories; thick solid curves are least-squares fitted circles (Kasa method); the green dashed circle is the theoretical radius. Inset confirms both policies sustain ${\sim}6.5\,\mathrm{m/s}$ throughout.
\textbf{(b)}~Turning radius comparison: heading-reward $24.2\,\mathrm{m}$ ($+30\%$) vs.\ yaw-rate $146.7\,\mathrm{m}$ ($+690\%$).
\textbf{(c)}~Per-lap center drift: heading-reward $1.7{\pm}0.9\,\mathrm{m}$ vs.\ yaw-rate $8.2{\pm}2.7\,\mathrm{m}$.
\textbf{(d,\,e)}~Cornering posture in Gazebo. The heading-reward policy leans $>13^\circ$ into the turn; the yaw-rate policy remains nearly upright.
}
\label{fig:turning_e2}
\end{figure}

Both policies use the full MHA actor and Multi-AMP discriminator; only the turning reward differs. On the turning test at $(6.5, 0, 0.35)$ (Figure~\ref{fig:turning_e2}):

\begin{itemize}[itemsep=0ex, leftmargin=1.2em,topsep=0.5ex]
  \item \textbf{Turning radius:} the heading-direction policy achieves $24.2\,\mathrm{m}$ ($+30\%$ vs.\ the theoretical $18.6\,\mathrm{m}$); the yaw-rate policy produces $146.7\,\mathrm{m}$ ($+690\%$)---approximately $6\times$ larger.
  \item \textbf{Trajectory consistency:} per-lap center drift is $1.7{\pm}0.9\,\mathrm{m}$ for the heading-direction policy vs.\ $8.2{\pm}2.7\,\mathrm{m}$ for the yaw-rate policy, indicating far better multi-lap repeatability.
  \item \textbf{Trunk lean:} the heading-direction policy exhibits pronounced inward trunk lean exceeding $13^\circ$ into the turn. The yaw-rate policy remains nearly upright.
\end{itemize}

These results confirm the analysis of Sec.~\ref{sec:method:yawhead}: the heading-direction reward lets the policy discover agile turning strategies that a yaw-rate reward actively penalizes.

\subsection{Effect of Per-Skill Style Supervision (E3)}
\label{sec:exp:e3}

\begin{table}[t]
\centering
\small
\setlength{\tabcolsep}{3.5pt}
\caption{E3 per-skill analysis in Isaac Lab (48 rollouts per skill). \textbf{Left:} DTW distance ($\downarrow$; mean per-frame L2 distance along the alignment, mean $\pm$ s.d.); \emph{Stand still} scores standing rollouts against each skill's reference~\citep{li2022wasabi}. \textbf{Right:} style score $\rho(D)$ of the in-place-turning discriminator of Multi-AMP ($D_{\mathrm{turn}}$) and of the shared discriminator (Sh.) on reference clips and on turning-command rollouts; compare within a row only, as Wasserstein scores are not calibrated across discriminators.}
\label{tab:dtw}
\begin{minipage}[t]{0.62\linewidth}
\centering
\begin{tabular}{lcccc}
\toprule
 & Stand & Turn & Walk & Run \\
\midrule
Multi-AMP & $\mathbf{0.23}{\scriptstyle\pm.02}$ & $\mathbf{2.62}{\scriptstyle\pm.01}$ & $\mathbf{2.72}{\scriptstyle\pm.04}$ & $\mathbf{4.64}{\scriptstyle\pm.03}$ \\
Shared-D & $0.25{\scriptstyle\pm.01}$ & $5.66{\scriptstyle\pm.04}$ & $3.79{\scriptstyle\pm.01}$ & $5.67{\scriptstyle\pm.07}$ \\
Stand still & -- & $5.76$ & $6.95$ & $14.34$ \\
\bottomrule
\end{tabular}
\end{minipage}\hfill
\begin{minipage}[t]{0.36\linewidth}
\centering
\begin{tabular}{lcccc}
\toprule
 & \multicolumn{2}{c}{Ref.} & \multicolumn{2}{c}{Rollout} \\
\cmidrule(lr){2-3}\cmidrule(lr){4-5}
$D$ & Stand & Turn & M. & Sh. \\
\midrule
$D_{\mathrm{turn}}$ & $0.49$ & $1.38$ & $1.21$ & $0.18$ \\
Sh. & $1.83$ & $1.56$ & $1.51$ & $0.45$ \\
\bottomrule
\end{tabular}
\end{minipage}
\end{table}

E3 replaces the four per-skill discriminators with a single discriminator trained on the pooled reference data; both runs sample the four skills with equal probability, which rules out data imbalance. We evaluate both policies in Isaac Lab under fixed commands for standing $(0,0,0)$, in-place turning $(0,0,0.6)$, walking $(1,0,0)$ and running $(4,0,0)$.

Both policies execute standing, walking and running, but \textbf{in-place turning collapses under the shared discriminator}. The Shared-D policy rotates at $0.08\,\mathrm{rad/s}$ instead of the commanded $0.6$ (Multi-AMP: $0.59$), keeps both feet on the ground $97\%$ of the time (Multi-AMP: $32\%$; reference: $33\%$) and lifts a foot $0.25$ times per second (Multi-AMP: $2.4$). Its turning DTW ($5.66$) is twice that of Multi-AMP ($2.62$) and matches that of standing still ($5.76$; Table~\ref{tab:dtw}): the turning skill has collapsed onto a quasi-static stance. Multi-AMP also attains lower DTW on walking ($-28\%$) and running ($-18\%$).

The cause is that \textbf{the shared discriminator is not skill-conditioned.} A single discriminator rates whether a transition resembles any reference motion, not whether it matches the commanded skill. It therefore scores standing and turning alike, both on the reference clips ($\rho{=}1.83$ vs.\ $1.56$) and on correctly executed rollouts ($1.44$--$1.52$ vs.\ $1.51$), whereas $D_{\mathrm{turn}}$ separates the two clips by $2.8\times$ (Table~\ref{tab:dtw}). Both discriminators score the collapsed motion lowest, so correct turning remains the optimum; what the shared discriminator lacks is the incentive to reach it. Under a turning command, its style term does not favor rotating over standing, and correct turning outscores not rotating only through the yaw-rate tracking reward, by about $0.03$ per step in units of $r_t$. With $D_{\mathrm{turn}}$, the style term raises this margin to $0.10$--$0.12$. The weak margin is insufficient for the policy to learn the stepping pattern that in-place turning requires; translating skills do not collapse because their velocity-tracking terms alone penalize standing still by about $0.08$ per step. Routing each transition to the discriminator of its commanded skill makes the style reward skill-conditioned, $D(\boldsymbol{\phi}\mid\skillvec)$, and restores the margin.

\subsection{Hardware Validation and Robustness}
\label{sec:exp:hardware}
To verify the achieved speed, we timed $100\,\mathrm{m}$ sprints on a standard athletics track (Figure~\ref{fig:hardware}). Under a $7.0\,\mathrm{m/s}$ command, a weight-reduced iteration of the full-scale platform (same actuators and kinematics) finished in $14\,\mathrm{s}$, a measured average of $7.14\,\mathrm{m/s}$; the deployment in Figure~\ref{fig:overview} runs stably at a $6.5\,\mathrm{m/s}$ command, but its ground speed was not measured. With the identical framework, a second humanoid ($1.36\,\mathrm{m}$, $40\,\mathrm{kg}$, $23$-DOF) finished in $22\,\mathrm{s}$ ($4.55\,\mathrm{m/s}$); its lower top speed reflects smaller actuators, as both robots reach their torque limits at top speed. Operating this close to the limits leaves little reserve for rejecting disturbances. In simulation, under velocity impulses at the maximum training magnitude ($\Delta v_x{=}{-}1.0\,\mathrm{m/s}$ backward, $\Delta v_y{=}1.0\,\mathrm{m/s}$ lateral; 50 trials each), the policy recovers from every push at $1.0$ and $5.5\,\mathrm{m/s}$, but only from $56\%$ of backward and $72\%$ of lateral pushes at $6.5\,\mathrm{m/s}$, consistent with the torque bottleneck in Sec.~\ref{sec:limitations}.

\section{Limitations}
\label{sec:limitations}

Training covers only mild terrain variation (rough, wavy and gently sloped ground); the purely proprioceptive policy would need exteroceptive sensing for stairs or unstructured ground. Transfer beyond our two kinematically similar humanoids remains untested.
The sustained top speed is bounded by joint torque rather than by the learning framework: at the highest hardware-tested speed, hip-pitch and knee-pitch joints operate near ${\sim}99\%$ of their torque limits and ankle-pitch near ${\sim}90\%$, and relaxing these limits in simulation raises the achievable speed. Beyond torque, an energy penalty is needed to keep joint power within motor capacity, and at still higher speeds the gait departs from the reference motion toward hardware-exploiting postures (lowered center of mass, exaggerated arm swing) that no motion prior guides, making suitable high-speed reference motions an open challenge.

\section{Conclusion}
\label{sec:conclusion}

We have shown that three targeted mechanisms---a causal multi-head attention history encoder, a heading-direction reward, and skill-conditioned multi-discriminator style supervision---can be integrated to train a single policy that sustains over $7\,\mathrm{m/s}$ sprinting and aggressive cornering with reference-consistent gait on a physical humanoid, deployed zero-shot from simulation. Each component addresses a distinct failure mode of high-speed locomotion: temporal attention stabilizes aggressive maneuvers, the heading-direction reward unlocks agile cornering with emergent trunk lean, and skill-conditioned discriminators keep every commanded skill from collapsing. Future work targets higher speeds and rough terrain with exteroception.

\clearpage

\bibliography{references}

\appendix
\counterwithin{figure}{section}
\counterwithin{table}{section}

\section{Timed Hardware Sprints}
\label{app:hardware}

\begin{figure}[h]
\centering
\includegraphics[width=0.9\columnwidth]{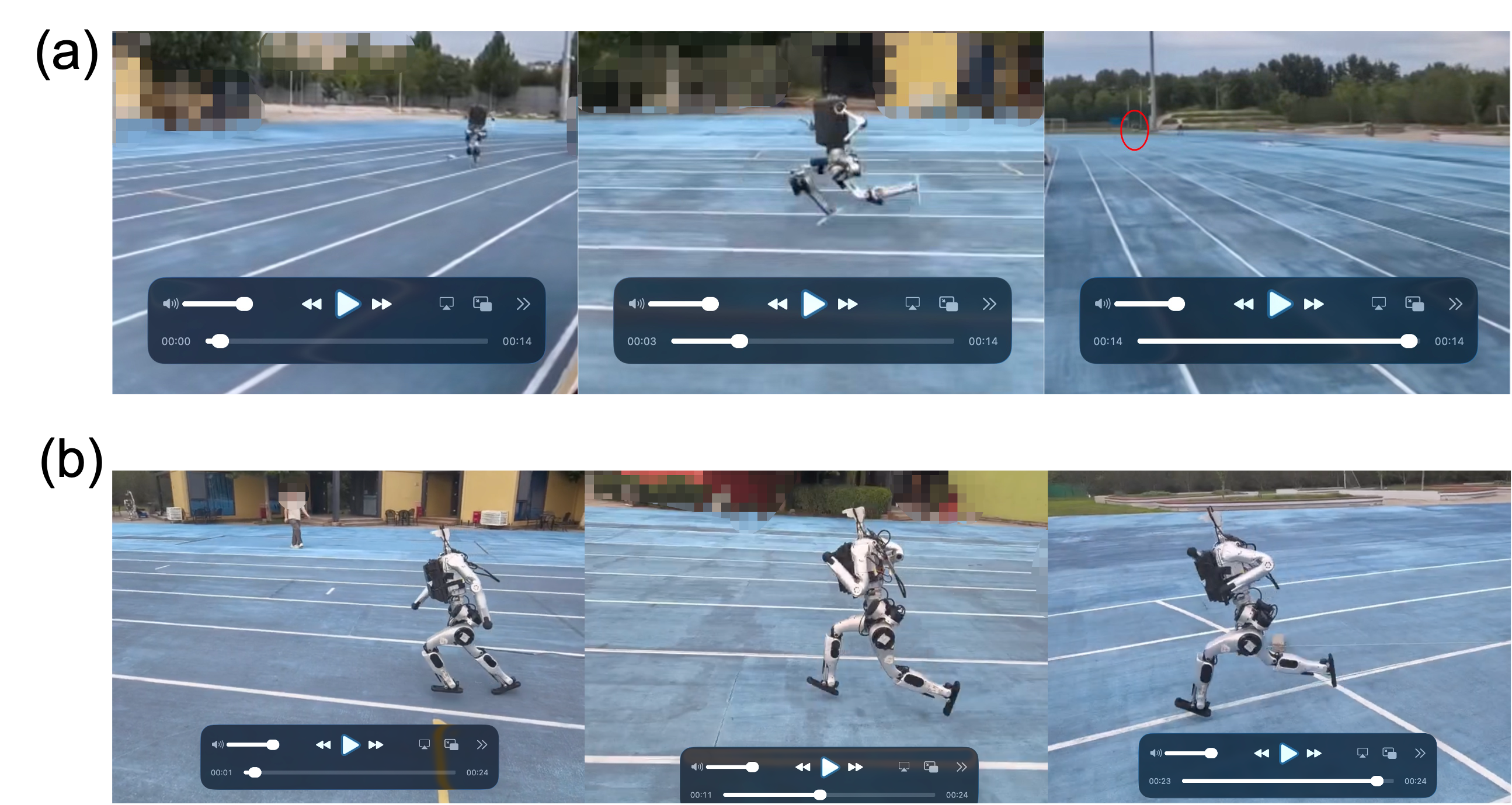}
\caption{%
Timed $100\,\mathrm{m}$ hardware sprints on a standard athletics track.
\textbf{Top:} primary platform (weight-reduced, same actuators and kinematics) completes $100\,\mathrm{m}$ in $14\,\mathrm{s}$ ($7.14\,\mathrm{m/s}$ avg.) under a $7.0\,\mathrm{m/s}$ velocity command.
\textbf{Bottom:} second humanoid ($1.36\,\mathrm{m}$, $40\,\mathrm{kg}$, $23$-DOF) completes $100\,\mathrm{m}$ in $22\,\mathrm{s}$ ($4.55\,\mathrm{m/s}$ avg.) using the identical algorithmic framework.
}
\label{fig:hardware}
\end{figure}

Figure~\ref{fig:hardware} shows timestamped video frames from the two timed $100\,\mathrm{m}$ sprints reported in Sec.~\ref{sec:exp:hardware}.

\section{Implementation Details}
\label{app:impl}

\textbf{Robot platform.} The primary platform is $1.8\,\mathrm{m}$ tall with $17$ actuated joints (Sec.~\ref{sec:exp:setup}). It weighs $70\,\mathrm{kg}$ in the configuration used for the simulation and deployment experiments and $60\,\mathrm{kg}$ in the weight-reduced iteration used for the timed sprint (Appendix~\ref{app:hardware}); the two share the same actuators and kinematics.

\textbf{State estimation.} The policy uses no learned or model-based state estimator. Its proprioceptive input (Sec.~\ref{sec:method:problem}) is read directly from onboard sensors: base angular velocity and roll/pitch angles from the torso-mounted IMU, and joint positions and velocities from the joint encoders. Base linear velocity is not observed by the actor; it is available only to the critic in simulation.

\textbf{Speed measurement.} The $100\,\mathrm{m}$ sprint distance was measured with a $100\,\mathrm{m}$ tape measure on a standard athletics track, and the completion time was obtained from video frame timestamps; the timestamped frames in Figure~\ref{fig:hardware} allow independent verification.

\textbf{Reference motions.} The four reference datasets (standing, in-place turning, walking, running) are human motion-capture clips retargeted to the robot with GMR~\citep{araujo2025gmr}, totaling $13{,}400$ frames ($268\,\mathrm{s}$) at $50\,\mathrm{Hz}$. Despite their unequal lengths, each skill is sampled with equal probability during discriminator training.

\textbf{Training compute.} Policies are trained on $8$ NVIDIA RTX~4090 GPUs with $4096$ parallel environments per GPU; a full training run of $10{,}000$ iterations takes about $15\,\mathrm{h}$.

\section{Hyperparameters}
\label{app:hyperparams}

Table~\ref{tab:hyperparams} lists all symbols used in the main text together with their numerical values. Table~\ref{tab:ppo} provides the full PPO and discriminator training configuration.

\begin{table}[h]
\centering
\caption{Symbols used in the main text and their numerical values.}
\label{tab:hyperparams}
\begin{tabular}{llr}
\toprule
Symbol & Description & Value \\
\midrule
\multicolumn{3}{l}{\textit{Robot}} \\
$h_{\mathrm{robot}}$ & Robot standing height & $1.8\,\mathrm{m}$ \\
\midrule
\multicolumn{3}{l}{\textit{Training (Sec.~3.5)}} \\
$f_{\mathrm{sim}}$ & Physics simulation frequency & $200\,\mathrm{Hz}$ \\
$n_{\mathrm{dec}}$ & Control decimation & $4$ \\
$f_{\mathrm{ctrl}}$ & Policy control frequency ($f_{\mathrm{sim}}/n_{\mathrm{dec}}$) & $50\,\mathrm{Hz}$ \\
$N_{\mathrm{env}}$ & Parallel environments & $8\times4096$ \\
$T_{\mathrm{ep}}$ & Episode length & $20\,\mathrm{s}$ \\
$K$ & Number of skills / discriminators & $4$ \\
$N_{\mathrm{ref}}$ & Total reference motion transitions & ${\sim}200{,}000$ \\
\midrule
\multicolumn{3}{l}{\textit{Discriminator (Sec.~3.4)}} \\
$\alpha$ & $\tanh$ output-bound scale & $0.3$ \\
$p$ & Gradient penalty exponent & $6$ \\
$\lambda$ & Task-style reward blend & $0.8$ \\
\midrule
\multicolumn{3}{l}{\textit{Heading-Direction Reward (Sec.~3.3)}} \\
$v_{\min}$ & Speed threshold for heading reward & $2.0\,\mathrm{m/s}$ \\
$\sigma$ & Exponential kernel width (Eq.~\ref{eq:heading_reward}) & $0.7$ \\
$\beta$ & L1 bias weight (Eq.~\ref{eq:heading_reward}) & $0.1$ \\
$k_\psi$ & Heading-error proportional gain (Eq.~\ref{eq:yawrate_cmd}) & $0.5$ \\
$\omega_{\min},\,\omega_{\max}$ & Yaw-rate clipping bounds (Eq.~\ref{eq:yawrate_cmd}) & $[-1.0,\,1.0]\,\mathrm{rad/s}$ \\
\midrule
\multicolumn{3}{l}{\textit{Evaluation}} \\
$v_{\max}$ & Step-test target speed & $6.5\,\mathrm{m/s}$ \\
\bottomrule
\end{tabular}
\end{table}

\begin{table}[h]
\centering
\caption{PPO and discriminator training hyperparameters.}
\label{tab:ppo}
\begin{tabular}{lr}
\toprule
Parameter & Value \\
\midrule
\multicolumn{2}{l}{\textit{PPO}} \\
Learning rate & $1\times10^{-3}$ (adaptive schedule) \\
Discount $\gamma$ & $0.99$ \\
GAE $\lambda$ & $0.95$ \\
Clip parameter & $0.2$ \\
Desired KL divergence & $0.01$ \\
Mini-batches per update & $4$ \\
Learning epochs per update & $5$ \\
Entropy coefficient & $0.005$ \\
Max gradient norm & $1.0$ \\
Value loss coefficient & $1.0$ \\
\midrule
\multicolumn{2}{l}{\textit{Discriminator}} \\
Learning rate & $5\times10^{-6}$ \\
Mini-batches per update & $80$ \\
Gradient penalty coefficient & $5$ \\
Loss function & Wasserstein \\
Replay buffer size & $200{,}000$ \\
Hidden layers & $[1024, 512]$ \\
Activation & LeakyReLU \\
AMP observation window & $4$ frames \\
\midrule
\multicolumn{2}{l}{\textit{Style Reward}} \\
Style reward coefficient ($w_s$) & $0.5$ \\
Task-style interpolation ($\lambda$) & $0.8$ \\
Reward mapping & $\rho(x)=\exp(\tanh(\alpha x))-e^{-1}$ \\
Symmetry data augmentation & Enabled \\
Reference state init ratio & $20\%$ \\
\bottomrule
\end{tabular}
\end{table}

\section{Architecture Details}
\label{app:arch}

\subsection{MHA Actor}

The MHA actor processes a $T{=}10$-frame proprioceptive history. Each per-step observation has dimension $D{=}63$: command (7) + base angular velocity (3) + base Euler angles (2) + joint positions (17) + joint velocities (17) + previous actions (17). The full architecture is listed in Table~\ref{tab:mha_arch}.

\begin{table}[h]
\centering
\caption{MHA actor architecture. Total actor-side parameters: $471{,}970$.}
\label{tab:mha_arch}
\begin{tabular}{llr}
\toprule
Component & Architecture & Params \\
\midrule
History embed & MLP: $56 \to 128 \to 128$ (ELU) & $23{,}808$ \\
Command embed & Linear: $7 \to 128$ & $1{,}024$ \\
Positional encoding & Sinusoidal (non-learnable) & $0$ \\
Causal Transformer & 1 layer, 2 heads, $d{=}128$, ff${=}256$, GELU, norm-first & $132{,}480$ \\
History output LN & LayerNorm($128$) & $256$ \\
Command fusion & Linear $256{\to}128$ + GELU + Linear $128{\to}128$ & $49{,}408$ \\
Fusion LN & LayerNorm($128$) & $256$ \\
Actor MLP & $191 \to 512 \to 256 \to 128 \to 17$ (ELU) & $264{,}721$ \\
Action std & Per-dimension scalar & $17$ \\
\bottomrule
\end{tabular}
\end{table}

The proprioceptive slice $\mathbf{p}_{t-k} \in \mathbb{R}^{56}$ is embedded via a shared two-layer MLP into a token $\mathbf{z}_{t-k} \in \mathbb{R}^{128}$, with sinusoidal positional encoding added. Tokens pass through a single pre-norm Transformer encoder layer with $h{=}2$ attention heads, feed-forward dimension $256$, GELU activation, and a causal mask so that $\mathbf{z}_{t-k}$ attends only to positions $\le t{-}k$. The output is layer-normalized and average-pooled across time to produce a dynamics summary $\bar{\mathbf{z}}_t \in \mathbb{R}^{128}$.

The current command $\mathbf{c}_t \in \mathbb{R}^7$ is embedded by a linear projection. The dynamics summary and command embedding are concatenated, passed through a two-layer MLP with GELU, and layer-normalized to produce $\mathbf{u}_t \in \mathbb{R}^{128}$. The actor head concatenates the current observation $\mathbf{o}_t$ with $\mathbf{u}_t$ and maps the result through a three-layer ELU MLP ($[512, 256, 128]$) to the action mean. Action variance is a learned per-dimension scalar.

The critic uses the privileged observation (including base linear velocity and foot contacts) through a separate MLP of the same hidden widths and is not Transformer-conditioned.

\subsection{MLP Actor (E1 Baseline)}

The E1 ablation replaces the entire MHA encoder with a plain MLP that directly consumes the flattened $T{\times}D{=}630$-dimensional history. Architecture: $630 \to 512 \to 256 \to 128 \to 17$ (ELU). Total actor-side parameters: $489{,}506$ ($3.6\%$ more than the MHA actor).

\subsection{AMP Observation Composition}

The AMP observation window $\boldsymbol{\phi}$ used by discriminators consists of a $4$-frame stack of the following per-frame features ($96$ dimensions per frame):

\begin{table}[h]
\centering
\caption{AMP observation composition per frame.}
\label{tab:amp_obs}
\begin{tabular}{lr}
\toprule
Feature & Dimension \\
\midrule
Joint positions & 17 \\
Key-point positions (selected body links) & 54 \\
Base linear velocity (XYZ) & 3 \\
Base angular velocity (RPY) & 3 \\
Joint velocities & 17 \\
Foot contact indicator & 2 \\
\midrule
\textbf{Total per frame} & \textbf{96} \\
\bottomrule
\end{tabular}
\end{table}

\subsection{Heading-Direction Reward Implementation Details}

The bias term $-\beta\,e_{\text{align}}$ in Eq.~(1) adds an L1 gradient outside the exponential's saturation regime, which is critical at the start of training when velocity-heading alignments are far from one.

During training, the commanded yaw rate $\omega^{\mathrm{cmd}}_z$ is not sampled independently but derived from the heading error via a proportional controller:
\begin{equation}
\omega^{\mathrm{cmd}}_z = \mathrm{clip}\!\big(k_\psi (\psi^{\star}_t - \psi_t),\; \omega_{\min},\; \omega_{\max}\big),
\end{equation}
so that the velocity command and heading target remain consistent. This coupling ensures the policy receives a coherent training signal: the heading reward drives the turning behavior while the commanded yaw rate provides a consistent velocity-command context.

\section{Sim-to-Real Bridge}
\label{app:sim2real}

\subsection{Domain Randomization}

Table~\ref{tab:dr} lists the full domain randomization ranges applied during training.
In addition to the dynamics randomization below, training terrain is procedurally generated with a difficulty curriculum, mixing uniform-noise rough ground ($50\%$, height noise $0.005$--$0.035\,\mathrm{m}$), wavy ground ($20\%$, amplitude $0.01$--$0.03\,\mathrm{m}$), and upward/downward pyramid slopes ($15\%$ each, grade $0.05$--$0.15$).

\begin{table}[h]
\centering
\caption{Domain randomization ranges.}
\label{tab:dr}
\begin{tabular}{llll}
\toprule
Parameter & Range & Operation & Timing \\
\midrule
Static / dynamic friction & $[0.3, 1.0]$ & Absolute & Startup \\
Restitution & $[0.0, 0.5]$ & Absolute & Startup \\
Joint default offset & $[-0.02, 0.02]\,\mathrm{rad}$ & Additive & Startup \\
Base COM offset (XYZ) & $\pm 0.05\,\mathrm{m}$ & Additive & Startup \\
Base mass perturbation & $\pm 5.0\,\mathrm{kg}$ & Additive & Startup \\
Actuator stiffness scale & $[0.8, 1.2]\times$ & Multiplicative & Each reset \\
Actuator damping scale & $[0.8, 1.2]\times$ & Multiplicative & Each reset \\
Joint friction scale & $[0.7, 1.5]\times$ & Multiplicative & Each reset \\
Joint armature scale & $[0.7, 1.5]\times$ & Multiplicative & Each reset \\
External push (linear) & $\pm 1.0\,\mathrm{m/s}$ (XY), $\pm 0.2$ (Z) & Impulse & Every $5$--$10\,\mathrm{s}$ \\
External push (angular) & $\pm 0.78\,\mathrm{rad/s}$ (RPY) & Impulse & Every $5$--$10\,\mathrm{s}$ \\
\bottomrule
\end{tabular}
\end{table}

\subsection{Actuator Model}

Each joint uses a \texttt{DelayedImplicitActuator} that models the first-order response of the real motors with a random actuation delay of $[0, 4]$ simulation steps ($0$--$20\,\mathrm{ms}$ at $200\,\mathrm{Hz}$). Per-joint stiffness and damping are derived from natural-frequency specifications: leg and waist joints use $2$--$5\,\mathrm{Hz}$ natural frequencies with damping ratios of $0.5$--$0.707$; arm joints use fixed stiffness ($50\,\mathrm{Nm/rad}$) and damping ($3\,\mathrm{Nm \cdot s/rad}$). Joint-level static and dynamic friction are modeled per joint group. Action scaling is computed as $0.25 \times \tau_{\max} / k_p$ per joint, where $\tau_{\max}$ is the effort limit and $k_p$ is the stiffness.

\subsection{Reward Function}

Table~\ref{tab:rewards} lists the complete task reward terms and their weights. The style reward from the Multi-AMP discriminators is blended with the task reward as described in Sec.~\ref{sec:method:camp}.

\begin{table}[h]
\centering
\caption{Task reward function terms and weights.}
\label{tab:rewards}
\begin{tabular}{llr}
\toprule
Term & Description & Weight \\
\midrule
\multicolumn{3}{l}{\textit{Velocity Tracking}} \\
track\_lin\_vel\_xy\_exp & XY velocity tracking (exponential, $\sigma{=}0.5$) & $5.0$ \\
track\_move\_dir\_to\_heading\_exp & Heading-direction alignment ($\sigma{=}0.7$) & $3.0$ \\
track\_lin\_x\_speed & Forward speed bonus & $2.0$ \\
\midrule
\multicolumn{3}{l}{\textit{Base Regulation}} \\
flat\_orientation\_exp & Torso orientation ($\sigma{=}0.5$) & $1.0$ \\
angular\_momentum\_exp & Angular momentum ($\sigma{=}3.0$) & $1.0$ \\
track\_robot\_height\_span\_exp & Target height $1.4\,\mathrm{m}$ & $1.0$ \\
\midrule
\multicolumn{3}{l}{\textit{Foot Regulation}} \\
feet\_slide & Foot sliding penalty & $-0.2$ \\
feet\_force & Excessive contact force & $-0.01$ \\
feet\_too\_near & Min foot separation ($0.2\,\mathrm{m}$) & $-5.0$ \\
\midrule
\multicolumn{3}{l}{\textit{Joint Regulation}} \\
joint\_deviation\_leg & Hip yaw / roll deviation & $-2.0$ \\
joint\_deviation\_upbody & Shoulder roll deviation & $-0.1$ \\
joint\_deviation\_standing & Arm / waist deviation (standing) & $-0.2$ \\
undesired\_contacts & Non-foot body contacts & $-1.0$ \\
\midrule
\multicolumn{3}{l}{\textit{Regularization}} \\
joint\_pos\_limits & Joint position limits & $-5.0$ \\
joint\_vel\_limits & Joint velocity limits (soft ratio $0.9$) & $-5.0$ \\
ang\_vel\_xy\_l2 & Base roll / pitch angular velocity & $-0.05$ \\
torque\_limits & Torque limit (ratio $0.9$) & $-0.01$ \\
hip\_pitch\_torque\_limits & Hip pitch torque (ratio $0.8$) & $-0.01$ \\
energy & Motor power squared & $-5\!\times\!10^{-3}$ \\
dof\_acc\_l2 & Joint acceleration & $-2.5\!\times\!10^{-7}$ \\
action\_rate\_l2 & Action rate penalty & $-0.1$ \\
\bottomrule
\end{tabular}
\end{table}

\section{Additional Results}
\label{app:results}

A video demonstrating all test protocols and ablation comparisons is available at \url{https://youtu.be/YWf62XkiApc}.

\end{document}